\def\year{2019}\relax
\documentclass[letterpaper]{article} %DO NOT CHANGE THIS
\usepackage{aaai19}  %Required
\usepackage{times}  %Required
\usepackage{helvet}  %Required
\usepackage{courier}  %Required
\usepackage{url}  %Required
\usepackage{graphicx}  %Required
\usepackage{svg}  %Required

\usepackage{amsfonts, bm} % formula
\usepackage{amsmath} % formula
\usepackage{float}
\usepackage{array}
\usepackage{booktabs} % tables
\usepackage[english]{babel}
\usepackage{multirow}
\usepackage{pdfpages,paralist}

\DeclareMathOperator{\GRU}{GRU}
\DeclareMathOperator{\BiGRU}{BiGRU}
\DeclareMathOperator{\softmax}{softmax}
\DeclareMathOperator{\head}{head}
\DeclareMathOperator{\MHead}{MultiHead}
\DeclareMathOperator{\Att}{Att}
\DeclareMathOperator{\embed}{embed}

\begin{document}
\title{Reactive Multi-Stage Feature Fusion for Multimodal Dialogue Modeling}
% \author{AAAI Press\\
% Association for the Advancement of Artificial Intelligence\\
% 2275 East Bayshore Road, Suite 160\\
% Palo Alto, California 94303\\
% }
%\author{Anonymous}
\author{
    Yi-Ting Yeh, Tzu-Chuan Lin, Hsiao-Hua Cheng, Yi-Hsuan Deng, Shang-Yu Su, Yun-Nung Chen\\
     National Taiwan University, Taiwan \\
     % \tt\small r07922064@ntu.edu.tw, 
     \tt\small \{r07922064,b04705003,b03902024,b04902013,f05921117\}@csie.ntu.edu.tw\quad y.v.chen@ieee.org
     % \tt\small b04902013@ntu.edu.tw \\
     % \tt\small shangyusu.tw@gmail.com
     % \tt\small yvchen@csie.ntu.edu.tw
 }

\maketitle
\begin{abstract}
% Introduce the structure of this paper
Visual question answering and visual dialogue tasks have been increasingly studied in the multimodal field towards more practical real-world scenarios.
%are difficult problems in multimodal learning field, while they are closer to real-life scenarios. 
A more challenging task, audio visual scene-aware dialogue (AVSD), is proposed to further advance the technologies that connect audio, vision, and language, which introduces temporal video information and dialogue interactions between a questioner and an answerer. 
This paper proposes an intuitive mechanism that fuses features and attention in multiple stages in order to well integrate multimodal features, and the results demonstrate its capability in the experiments.
Also, we apply several state-of-the-art models in other tasks to the AVSD task, and further analyze their generalization across different tasks.

\end{abstract}

\section{Introduction}
% simply introduce multi-modality learning, visual question answering and visual dialog(only static image).

Nowadays, many real-world problems require to integrate information from multiple sources.
Sometimes such problems involve distinct modalities, such as vision, language and audio.
Therefore, many AI researchers are trying to build models that can deal with tasks using different modalities, and multimodal applications have become an increasing popular research topic.
For example, visual question answering (VQA) \cite{vqa} is the task that aims at testing whether an AI model can successfully summarize an image, text description of that image and then generate a correct response.
The visual dialog task focuses on examining whether a system can interact with human via conversations~\cite{visual_dialog}, where in a conversation, the system is expected to answer questions correctly given an input image.
However, it is difficult to well converse with users when only accessing a single image without audio and dynamic scenes.
%The further extension - visual dialogue task proposed by \cite{visual_dialog}, wants to challenge an AI model to interact with a human by a conversation.
%In a conversation, an AI model must answer turns of human's questions correctly given a input image. However, to have a conversation with user about what is going on in the user's surroundings, these tasks which only consider single image but no audio and dynamic scenes, have their restrictions.

Motivated by the need for scene-aware dialogue systems, audio visual scene-aware dialogue (AVSD) is recently proposed by \citeauthor{avsd_dataset}, providing more dynamic scenes with video modal rather than static images in VQA or Visual dialogue tasks. With richer information, machines can be trained to carry on a conversation with users about objects and events around them.
To tackle this problem, this paper focuses on 1) better encoding multimodal features and 2) better decoding the responses with consideration of the encoded information.
We propose an reactive encoder that is capable of fusing multimodal features and paying attention correctly, and a decoder that investigates different decoding mechanisms for better generating the dialogue responses.
% What we do, from feature selection (multi-modality) to architecture search (question answering)?
In this paper, our main contributions are 3-fold: 
\begin{compactitem}
    \item This paper proposes a simple but effective way to fuse different modalities by the 1 \(\times\) 1 convolution followed by a weighted sum operation.
    \item The proposed multi-stage fusion mechanism can encourage the model to thoroughly understand the question.
    \item This paper first attempts to generalize top-down attention proposed by \citeauthor{anderson2018bottom} and investigates different attentional decoding methods in a principled way.
    %\item Propose a post-processing method that can help the model generate more accurate responses.
\end{compactitem}

\section{Task Description}
\begin{figure*}[t]
    \centering
    \includegraphics[width=\linewidth]{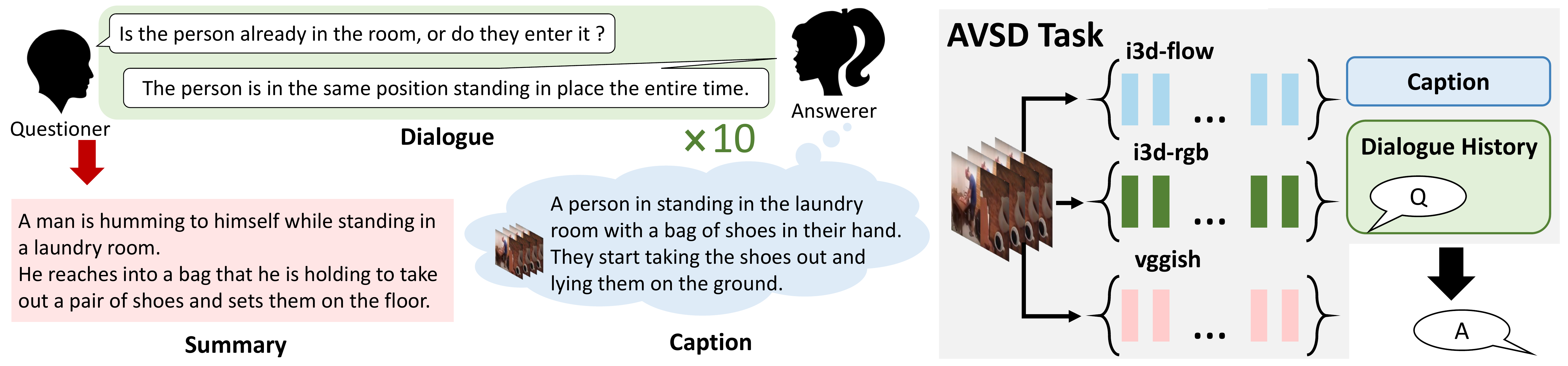}
    \caption{The illustration of the collected data and the audio visual scene-aware dialogue (AVSD) task.}
    \label{fig:data_collect}
\end{figure*}

% introduce dataset
The task is audio visual scene-aware dialogue (AVSD), which tests the capability of dialogue responses given audio, visual and dialogue contexts.
The process of data collection is illustrated in the left part of Figure \ref{fig:data_collect}, where the collected dataset contains 11,156 visual dialogues.
Each dialogue contains a video and dialogue texts, where there are pre-extracted feature modalities using VGGish \cite{hershey2017cnn} (\textbf{vggish}) and I3D models \cite{carreira2017quo} (\textbf{i3d}) for the video, and the dialogue texts contain a video caption, a summary written by the questioner after 10 rounds Q/A and dialogue history, denoted as \textbf{caption}, \textbf{summary} and \textbf{dialogue} respectively.
Note that our model in the experiments only considers \emph{answer} parts as our \textbf{dialogue} features.
%ne thing to mention is that we think questions in the dialogue history are indefinite information and may mislead our model, so we remove questions and only keep answers in the \textbf{dialog} feature.
Our goal is to play the role of an answerer, who can reply the reasonable answer to the questioner illustrated in the right part of Figure~\ref{fig:data_collect}.

\iffalse
\begin{figure}[t]
    \centering
    \includegraphics[width=\linewidth]{figures/taskDes.png}
    \caption{Audio Visual Scene-Aware dialogue Dataset Example}
    \label{fig:avsd}
\end{figure}
\fi

\section{Proposed Approach}

Our system can be viewed as an encoder-decoder model, which is commonly used in conversation modeling \cite{sutskever2014sequence}. We first explain the feature selection procedure, describe the basic fusion model, and then detail the novel design for the proposed model.
In the encoder, we apply basic fusion, multi-stage fusion, and 1x1 convolution fusion methods.
In the decoder, we apply an original decoder, an attention decoder, and a top-down attention LSTM decoder. 
The detail is described below.

\subsection{Feature Selection}

% feature definitions, why do not use summary and i3d. choose some example in official testset to show summary sucks.
Considering that multiple modalities are involved in this task, it is important to choose proper and useful features for it.
%In this section we will describe provided features at first, and then explain why we selected specific features.
%For each video, audio and visual information is extracted using VGGish model \cite{hershey2017cnn} and I3D model \cite{carreira2017quo} respectively, and we will call these features \textbf{vggish}, \textbf{i3d} \footnote{In fact, \textbf{i3d} feature has two parts \textbf{i3d-flow} and \textbf{i3d-rgb}, but for simplicity we call them \textbf{i3d} and always use them together} in the rest of paper. Besides video, we are provided with video caption, summary written by questioners after 10 rounds Q/A and previous dialogue history to answer a question, denoted as \textbf{caption}, \textbf{summary}, and \textbf{dialog}. Because we think questions in the dialogue history are indefinite information and may mislead our model, we remove questions and only keep answers in the \textbf{dialog} feature. 
However, some features are too noisy to model and may result in ambiguity.
In our proposed model, we decide to ignore \textbf{i3d} and \textbf{summary} for our feature set.

There are several reasons to drop \textbf{i3d}.
The main reason is that \textbf{i3d} has too much information and thus contains many noises, which may hurt the stability of our model.
We assume that the useful information contained in \textbf{i3d} can be obtained by combining \textbf{vggish} and texts.
Because the audio tends to attract attention of people and be the key in the video, \textbf{vggish} provides more compact information and texts supply detailed description. 
In our experiments, we also found that removing \textbf{i3d} does not lead to a huge degradation of performance, implying that in fact we can answer most questions without this feature set in our model setup. 

The reason of removing \textbf{summary} in our feature set is the concern about misleading and impracticality.
We observe that \textbf{caption} usually contains more concise description and more rich information than \textbf{summary}, so our model only concentrates on \textbf{caption} to better answer questions.
Another consideration is about how \textbf{summary} is generated.
It is also more reasonable and practical to remove \textbf{summary} from our feature set, considering that the summary is generated by the questioner who does not see the video.
For example, Table \ref{tab:mislead_summary} shows a misleading \textbf{summary} example that contains incorrect information.

% example in testset4DSTC XBG8W
\begin{table}[t!]
    \centering
    \small
    \begin{tabular}{ l p{6cm} }
        \hline
        \bf Type & \bf Sentence\\ \hline\hline
        \textbf{Caption} & Man walks over to laptop and throws towel over his shoulder. He sits down and wipes and scratches his face with his hands and begins staring at the laptop. \\ \hline
    \textbf{Summary} & A man walks into a room and sits down while looking at a laptop on the \textbf{\textit{floor}}. \\ \hline
    \end{tabular}
    \vspace{-2mm}
    \caption{Misleading \textbf{summary} example.}
    \label{tab:mislead_summary}
\end{table}

\begin{figure*}[t!]
    \centering
    %\includesvg[width=\linewidth]{figures/model.svg}
    \includegraphics[width=0.87\linewidth]{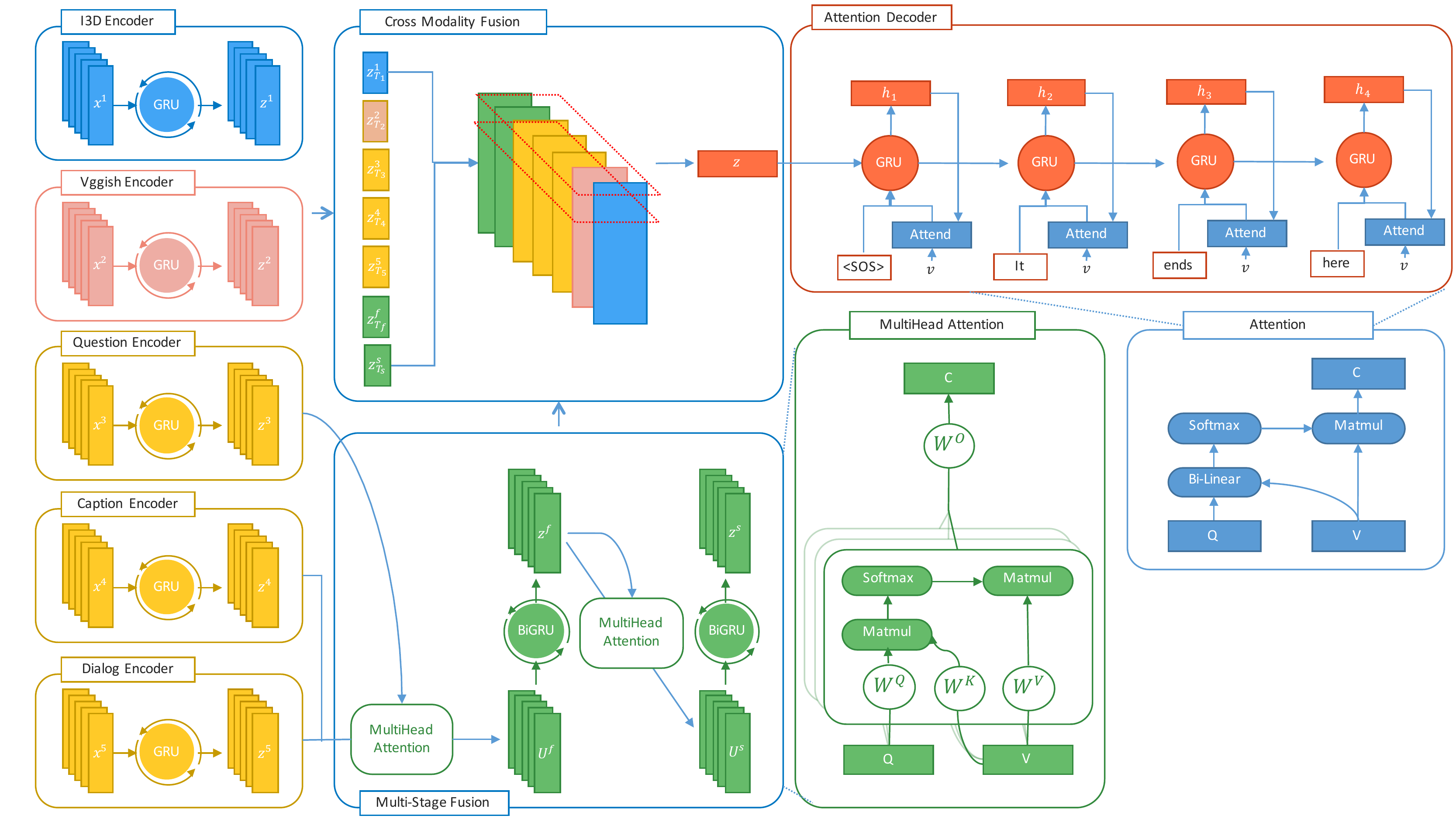}
    \vspace{-2mm}
    \caption{The illustraion of the proposed model architecture.}
    \vspace{-2mm}
    \label{fig:model}
\end{figure*}

\subsection{Feature Encoder} 
After removing noisy features, our model utilizes several sequential features in different forms for the task. 
Let $\bm{x}^i = \{\bm{x}^i_1, \dots \bm{x}^i_{T_i}\}$ represent the $i$-th feature, which can contain both video features and text word embeddings.
For text-based features, \emph{Glove} is applied to obtain word embeddings.
Then for each temporal sequence of features, an uni-directional GRU encoder is used to encode all sequential features into a common space $\mathbb{R}^{d}$, producing $\bm{z}^i = \{\bm{z}^i_1, \dots \bm{z}^i_{T_i}\}$, where $\bm{z}^i_t \in \mathbb{R}^{d}$ and can be the representation of the whole video or the dialogue history.
The illustration can be found in Figure~\ref{fig:data_collect}.

\subsection{Basic Fusion Model} 
In our basic fusion model, we use the last hidden states $\{\bm{z}^1_{T_1}, \dots, \bm{z}^n_{T_n}\}$ for each modality as the encoded representation, and fuse different modalities into a global multimodal representation, $\bm{z}$, using a weighted sum operation.
$T_i$ indicates the number of hidden states of the i-th feature. The fused representation is computed by
\begin{equation}
\label{eq:1x1conv}
\bm{z} = \frac{\sum^n_{i=1} \bm{w}_i \bm{z}^i_{T_i}}{\sum^n_{i=1} \bm{w}_i}, 
\end{equation}
where $\bm{w}_i$ is a trainable scalar weight and $n$ is the number of feature types.

We initialize the first hidden state $h_0 = \bm{z}$ in the decoder and use a GRU to generate the answer.
The decoding procedure follows $h_{t+1} = \GRU(y_{t}, h_{t})$, where $h_{t}$ is the $t$-th hidden state in the decoder and $y_t$ is the output in the $t$-th step.
\begin{equation}
\label{eq:decode}
\Pr(y_{t} \mid y_{1:t-1}) = \softmax(Wh_{t} + b),
\end{equation}
where $W \in \mathbb{R}^{|\Sigma| \times d_h}$ and $b \in \mathbb{R}^{|\Sigma|}$ are learned weights. $\Sigma$ and $d_h$ are the vocabulary and the dimension of the hidden state respectively.

\subsection{Multi-Stage Fusion}

In addition to the above basic fusion model, here we introduce a novel feature fusion mechanism using multi-stage attention.
The attention enables the model to focus on a targeted area within a context that is relevant to answer the question \cite{vqa,xiong2016dynamic}.
With the attention mechanism, we can enhance or modify vectors set with the information from other vectors; it is called the fusion process where information of some vectors is fused into other vectors \cite{huang2018fusionnet}.

An general attention function can be described as computing the weighted sum of values, where the weight is given by the attention score between a query and a key via a score function.
For simplicity, our model uses the same vector for the value and the key.
Here we choose the recently popular \emph{multi-head attention} as our attention score function \cite{vaswani2017attention}:
\begin{equation}
\label{eq:multihead}
\begin{aligned}
\MHead(Q, V) &= [\head_1; \dots ;\head_n] W^O,\\
\head_i &= \Att(QW^{Q}_i, VW^{K}_i, VW^{V}_i),\\
\Att(Q, V, K) &= \softmax(\frac{QV^{T}}{\sqrt{d_k}})K, 
\end{aligned}
\end{equation}
where $W^{Q}_i \in \mathbb{R}^{d \times d_k}$, $W^{K}_i \in \mathbb{R}^{d \times d_k}$, $W^{V}_i \in \mathbb{R}^{d \times d_k}$ and $W^O \in \mathbb{R}^{nd_k \times d}$ are parameters.
$[A;B]$ means the concatenation of two vectors $A$ and $B$.
$n$ is the number of heads and $d$ is the dimension.
$d_k = d / n$. Note that we add a residual connection after the attention to help model converge faster~\cite{he2016deep}.

Our multi-stage fusion starts from fusing encoded \textbf{question} ($z^Q$) into encoded \textbf{caption} ($z^C$) and \textbf{dialogue} ($z^D$), and we consider the concatenation of \textbf{caption} and \textbf{dialogue} to be the context.
We apply the multi-head attention as (\ref{eq:multihead}) to generate fusion vectors and then feed them into a bidirectional GRU reader to obtain the fused representation:
\begin{equation}
\label{eq:mheadfusion}
\begin{aligned}
U^{fusion} &=  \MHead([z^C;z^D], z^Q) \\
z^{fusion} &= \BiGRU(U^{fusion}) \\
\end{aligned}
\end{equation}

While the context is necessary to infer the answer, an issue in our current fusion representation is lack of full understanding about the context. 
To address this problem, we compute the attention $U^{self}$ using \emph{self-attention}, which usually improves performance a lot in machine comprehension and question answering \cite{P17-1018,vaswani2017attention}.
We also feed $U^{self}$ into a bidirectional GRU and obtain $z^{self}$, where $z^{self}$ can be viewed as vectors that fuse all textual information together. Formally written as:
\begin{equation}
\label{eq:mheadself}
\begin{aligned}
U^{self} &= \MHead(z^{fusion}, z^{fusion}), \\
z^{self} &= \BiGRU(U^{self}). \\
\end{aligned}
\end{equation}
Then we have tried many different ways to leverage $z^{fusion}$ and $z^{self}$, and find the most efficient way is to use the last hidden state of them and add them into computation of the weighted sum in \eqref{eq:1x1conv}.
Therefore, the multi-stage fusion mechanism intuitively offers richer information to enhance the first hidden state passed to the decoder.

\subsection{1x1 Convolution Fusion}

Instead of a simple weighted sum in \eqref{eq:1x1conv}, we find that it is beneficial to insert 1x1 convolution \cite{szegedy2015going} before the weighted sum to help each feature channel interact with each other.
The weighted sum operation can also be considered as an single channel 1x1 convolution with output normalization.

To model the multimodal features using $n$ channels, we utilize the last hidden state for each feature type, $\{\bm{z}^1_{T_1}, \dots, \bm{z}^n_{T_n}\}$.
We run $n_c$ channels 1x1 convolution on the inputs and obtain $\{\hat{z}^1, \dots, \hat{z}^{n_c}\}$.
Then we compute the weighted sum as \eqref{eq:1x1conv}: 
\begin{equation}
\label{eq:multichannel_1x1conv}
\bm{z} = \frac{\sum^{n_c}_i \bm{w}_i \hat{z}^i}{\sum^{n_c}_i \bm{w}_i},
\end{equation}
where $\bm{w}_i$ is a trainable scalar weight.

\subsection{Attention Decoder}

In the decoding phase, attention is also commonly used to help model focus on important information in the encoder-decoder model \cite{Luong2015EffectiveAT}.
We choose a commonly used multiplicative attention \cite{Britz:2017} and the computation follows:
\begin{equation}
\begin{aligned}
a_{i,t} &= v^{T}_{i}U^{T}V h_{t}, \\
\alpha_t &= \softmax(\bm{a}_{t}),
\end{aligned}
\end{equation}
where $U \in \mathbb{R}^{k \times d_x}$, $V \in \mathbb{R}^{k \times d_y}$ are trainable weights and $v_{i} \in \mathbb{R}^{d_x \times 1}$, $h_{t} \in \mathbb{R}^{d_y \times 1}$ are inputs.
$k$ is the attention dimension. 
Note that $a_{i, t} = v^{T}_{i}U^{T}Vh_{t}$ is also called low-rank bilinear method \cite{pirsiavash2009bilinear,kim2016hadamard}.

In the decoding step $t$, the attention of a query $h_{t}$ and values $\bm{v} = \{\bm{v}_1, \dots \bm{v}_r\}$ is computed as:
\begin{equation}
C_{t} = \sum^r_i \alpha_{i, t} \bm{v}_i.
\end{equation}
We then concatenate the context vector $C_{t}$ with the next step input $y_{t}$ as the attention-enhanced input and pass it through GRU, formally written as:
\begin{equation}
h_{t+1} = \GRU([y_{t};C_{t}], h_{t}).    
\end{equation}

Here we choose a simple concatenation of all encoded features to form $\bm{v}$: $\{\bm{z}^1_1, \dots, \bm{z}^1_{T_1}, \bm{z}^2_1, \dots, \bm{z}^{n}_{T_n}\}$.
We will discuss how $\bm{v}$ influences performance in the experiments.

\subsection{Top-Down Attention LSTM}

We also try an alternative attention decoder called \emph{top-down attention LSTM} \cite{anderson2018bottom}, which has been proved useful in visual question answering. The original design makes this model selectively attend to spatial image features, but we generalize it and feed arbitrary encoded values set $\bm{v} = \{\bm{v}_1, \dots, \bm{v}_r\} = \{\bm{z}^1_1, \dots, \bm{z}^1_{T_1}, \bm{z}^2_1, \dots, \bm{z}^{n}_{T_n}\}$ into this model.
We consider this model as an enhanced attention decoder, where an additional attention LSTM is used to further track what has been attended and expect it can provide more useful information for the attention module.

%\begin{figure}[t]
%     \begin{center}
%         \includegraphics[width=0.95\linewidth]{figures/Top_Down_Attn_fit.png}
%     \end{center}
%     \caption{Overview of the top-down attention LSTM. Two LSTM layers are used to selectively attend to features sequences $\{\bm{h}_1, \dots  \bm{h}_i\}$. These features in our model is concatenation of the GRU encoded input features}
%     \label{fig:captioner}
% \end{figure}

Here the top-down attention LSTM has two LSTM layers, which are named \emph{attention LSTM} and \emph{language LSTM}.
We consider the first LSTM layer as an attention model and the second LSTM layer as a language model.
The main purpose of separating two LSTM cells is to modularize the learning process into two parts: capturing the importance of multimodal features (attention LSTM) and language generation (language LSTM). 

The input vector to the attention LSTM at time step $t$ consists of the previous output hidden states of the language LSTM concatenated with the mean pooled features $\bar{v} = \frac{1}{r} \sum^r_{i=1} v_i$ and a word embedding of the previously generated word.
$\bm{x}^1_t = [h^2_{t-1};\bar{v};\embed(y_{t-1})]$, where $\embed(y_{t-1})$ is the word embedding of $y_{t-1}$.
Given the output $h^1_t$ of the attention LSTM, we compute the attention score as follows:
\begin{equation}
\begin{aligned}
b_{i, t} &= w^T_{a}\tanh(W_{va}\bm{v}_i + W_{ha}h^1_t), \\
\beta_t &= \softmax(\bm{b}_{t}).
\end{aligned}
\end{equation}
Then we calculate the context vector of attended features as 
\begin{equation}
C_{t} = \sum^r_i \beta_{i, t} \bm{v}_i.
\end{equation}

The input to the language LSTM consists of the context vector of attended vectors concatenated with the output of the attention LSTM, $\bm{x}^2_t = [h^1_t;C_{t}]$.
We pass $\bm{x}^2_t$ into the language LSTM to obain $h^2_t$ and compute the output word distribution as:
\begin{equation}
\Pr(y_{t} \mid y_{1:t-1}) = \softmax(Wh^2_{t} + b),
\end{equation}
which is exactly same as the basic decoding setup in \eqref{eq:decode}.

\subsection{Training and Testing}
The full model is trained in an end-to-end manner to optimize the dialogue generation results in \eqref{eq:decode}.
During testing, beam search is applied to generate the fluent answering sentences given the contexts.

\begin{table*}[t!]
    \begin{center}
    \small
    \begin{tabular}{lll|ccccccc}
    \toprule
        & {Model Encoder} & Model Decoder & {B-1} & {B-2} & {B-3} & {B-4} & {MET.} & {R-L} & {CIDEr}\\
    \midrule
         Baseline & \multicolumn{2}{l|}{Na\"{i}ve Copy} & 0.231&0.124&0.077&0.049&0.111&0.235&0.637\\
         & \multicolumn{2}{l|}{Released \cite{alamri2018audio}} & 0.270 & 0.172 & 0.118 & 0.085 & 0.115 & 0.292 & 0.790\\
         \midrule
         Basic Fusion & Simple Fusion & Simple & 0.232 & 0.157&0.112&0.084&0.120&0.305&0.994\\
         & Multi-Stage  & Simple & 0.231 & 0.157 & 0.113 & 0.086 & 0.119&0.308&1.009\\
         & 1x1 Convolution & Simple & 0.239&0.162&0.117&0.088&0.122&0.310&1.013\\
         \cmidrule{2-10}
         & Simple Fusion & Attention & \textbf{0.245}&0.162&0.115&0.086&0.119&0.308&0.977\\
         & Simple Fusion & Top-Down Attention LSTM  & 0.234 &0.158&0.113&0.085&0.119&0.309&0.986\\
         \midrule
         Multi-Stage  & Multi-Stage & Attention  & 0.238&0.161&0.116&0.088&0.123&0.311&1.028\\
         & Multi-Stage + 1x1 Conv & Attention & 0.238&0.163&0.118&0.090&0.122&\textbf{0.315}&\textbf{1.059}\\
         & Multi-Stage + 1x1 Conv & Attention (w/o \textbf{vggish}) & 0.243&\textbf{0.165}&\textbf{0.119}&\textbf{0.091}&\textbf{0.124}&0.313&1.046\\
         & Multi-Stage + 1x1 Conv & Attention (w/ \textbf{i3d}) &
         0.234&0.157&0.112&0.084&0.117&0.303&0.958\\
         \midrule
         Submitted & \multicolumn{2}{l|}{Basic Fusion Model (Prototype)} & 0.237&0.161&0.116&0.088&0.121&0.310&1.015\\
         & \multicolumn{2}{l|}{+ Fine-Grained Response Revision}& 0.238&0.161&0.116&0.087&0.122&0.315&1.024\\
         & \multicolumn{2}{l|}{Fusion Text Model}& 0.214&0.140&0.098&0.073&0.110&0.286&0.859\\
         & \multicolumn{2}{l|}{+ Fine-Grained Response Revision}& 0.215&0.141&0.099&0.073&0.111&0.291&0.874\\
         
    \bottomrule
    \end{tabular}
    \end{center}
    \vspace{-3mm}
    \caption{\label{tab:results} The results of baselines and our proposed models, where our models do not use \textbf{i3d} and \textbf{summary}. %$^\ast$ means model does not use \textbf{vggish} i.e. the pure text model. $^\Psi$ means model also use \textbf{i3d} features.
    }
    \vspace{-2mm}
    \end{table*}

\section{Experiments}
To evaluate whether the proposed model is capable of modeling dialogues with multimodal scenarios, a set of experiments is conducted.

\subsection{Experimental Setup}

We run our experiments using the official AVSD dataset \cite{avsd_dataset}, which consists of 7659, 1787, 1710 dialogues for train, dev, and test sets respectively.
In the dataset, almost all dialogues contain 10 question/answer pairs.
We use the evaluation script \texttt{nlg-eval} \footnote{https://github.com/Maluuba/nlg-eval} to compute objective evaluation scores for the model, where the punctuations, `,' and `.', are not taken into account.
In the experiments, the compared baselines include a na\"{i}ve copy baseline and the released baseline~\cite{alamri2018audio}, where the na\"{i}ve baseline simply copies the questions as its answer and serves as a lower bound of this task.
%Besides our model, we also report a naive copy baseline, which simply copy the question as its answer , and released baseline \cite{hori2018end}.
Note that because we found there are many word overlaps between the question and the answer, the copy baseline in fact has very strong performance under objective evaluation metrics. 
%We evaluate released baseline with nlg-eval too.
%During objective evaluation phase, we remove punctuation "," and "." from human references and our model output.

\subsection{Fine-Grained Response Revision}
In the answering sentence to the yes/no question, adding ``yes'' or ``no'' at the beginning may affect the performance.
%We observe that whether there is "yes" or "no" at the beginning of a sentence affects the performance of automated metrics.
Therefore, we train a classifier to predict whether ``yes'' or ``no'' tokens should be inserted.
This classifier consists of an RNN encoder followed by a linear projection output layer.  
Note that we first filter out yes/no questions by rules, since yes/no questions should not be answered with this token.

\subsection{Results}

Table~\ref{tab:results} shows the experimental results, where our models only take \textbf{vggish}, \textbf{caption}, and \textbf{dialog} into account, while the released baseline additionally use \textbf{i3d} as their input features. 
While the basic model is the simple architecture without any mechanism, it shows surprising performance specific in CIDEr (from 0.69 to 0.99) \cite{Vedantam2015CIDErCI}. 
Because CIDEr measures the ability of capturing correct objects in the image and places less attention on words which frequently appear in answers such as stopwords, the result tells that the basic model already captures important information and objects in the video without the complicated design of attention. 
In the proposed model, we analyze the results for different encoding and decoding mechanisms.

\paragraph{Encoding}
We examine the effectiveness of the proposed two fusion methods, multi-stage fusion and 1x1 convolution fusion.
The proposed multi-stage fusion provides model deeper understanding of input features and then further boosts the performance of CIDEr.
The 1x1 convolution fusion also enable interactions between different modalities and improves the performance.
\paragraph{Decoding}
Furthermore, two decoding mechanisms, attention decoder and top-down attention LSTM, are investigated.
Two variants of decoders have different advantages.
The attention decoder mainly improves BLEU scores \cite{papineni2002bleu}, while the top-down attention boosts the performance for the ROUGE score \cite{lin2004rouge}.
It may be because the attention decoder enhances the ability of our model to capture the specific word usage in dialogues.
On the other hand, the top-down attention LSTM adds the attention LSTM to help track what is attended in previous decoding, and thus improves the ROUGE-L where the recall of generated words is relatively important.
Hence, the attention decoder is applied as our decoder in the following experiments due to its strong
%The reason to choose Attention Decoder in full model but not Top-Down Attention LSTM is because we want to incorporate the strong
ability of learning how to use particular word in certain context and think these modules will have complementary effects.

\paragraph{Proposed Fusion Model}
Our full fusion model is composed of multi-stage fusion, 1x1 convolution fusion, and the attention decoder, and further improvement is investigated and analyzed in the experiments shown in Table~\ref{tab:results}.
Compared to the released baseline, we can find that even the proposed basic model can outperform it by a large margin in CIDEr, improve METEOR and ROUGE-L, and achieve comparable results in BLEU-4.
Note that BLEU-1 and BLEU-2 are not good metrics to evaluate quality of generated sentences in dialogues, because they only measure unigram and bigram precision. 
The copy baseline shows that we can reach high BLEU-1 and BLEU-2 by simply copying the questions, implying that these are not good indicators.
The comparable BLEU-4 shows that our basic model in fact have similar ability to capture the sentence structure compared to the released baseline.
Furthermore, our proposed full fusion model can significantly outperforms the released baseline for almost all metrics.

Here we also conduct experiments on two variants of our full fusion model.
The first one does not use \textbf{vggish} features, so it is a pure text model, and the second one additionally uses \textbf{i3d} features, which can be fairly compared with the released baseline.
Table~\ref{tab:results} shows that the pure text model also achieves better BLEU scores compared to the model that uses additional video features, implying that the language part of test data provides rich enough information for the model to answer most questions.
On the other hand, using \textbf{i3d} features results in performance drop, confirming our concern about noisy \textbf{i3d} features and demonstrating the need of the proposed feature selection.

\begin{table}[t!]
    \begin{center}
    \small
    \begin{tabular}{l|ccccccc}
    \toprule
        {Method} & {B-1} & {B-2} & {B-3} & {B-4} & {MET.} & {R-L} & {CIDEr}\\
    \midrule
         Prod & .221 & .146 & .103 & .076 & .109 & .288 & .823\\
         Sum & .232&.156&.111&.083&.116&.305&.950\\
         Concat & \textbf{.237} & \textbf{.160} & \textbf{.116} & \textbf{.087} & \textbf{.120} & \textbf{.306} & .987\\ 
         Weight & .232&.157&.112&.084&\textbf{.120}&.305&\textbf{.994}\\
    \bottomrule
    \end{tabular}
    \end{center}
    \vspace{-3mm}
    \caption{\label{tab:fusion_methods} Results of different fusion methods. }
    \end{table}

    \begin{table*}[t!]
    \begin{center}
    \small
    \begin{tabular}{l|cccccccc}
    \toprule
        {Model} & {BLEU-1} & {BLEU-2} & {BLEU-3} & {BLEU-4} & {METEOR} & {ROUGE-L} & {CIDEr} & {Human}\\
    \midrule
         Baseline (i3d) \cite{alamri2018audio} &
        .621& .48& .379& .305& .217& .481& .733&---\\
        Baseline (i3d + vggish) \cite{alamri2018audio} &
        .626& .485& .383& .309& .215& .487& .746& 2.848\\
        \midrule
        Basic Fusion Model (Official) &
         .636& .510&\textbf{.417}&\textbf{.345}&\textbf{.224}&\textbf{.505}&\textbf{.877}&---\\
         Basic Fusion Model (Prototype) &
         .640&\textbf{.513}& .416& .342& .223& .504& .837& \textbf{3.188}\\
         + Fine-Grained Response Revision  & 
         \textbf{.641}&\textbf{.513}& .416& .342& .223& .504& .836&---\\
         Fusion Text Model &
         .592& .468& .375& .304& .206& .475& .729& 2.928\\ 
         + Fine-Grained Response Revision &
         .595& .477& .376& .304& .207& .477& .731&---\\
    \bottomrule
    \end{tabular}
    \end{center}
    \vspace{-2mm}
    \caption{\label{tab:official_results} The submitted results in the official test set.  }
    \vspace{-2mm}
    \end{table*}
   
\begin{table}[t!]
    \begin{center}
    \small
    \begin{tabular}{l|ccccccc}
    \toprule
         {Features} & {B-1} & {B-2} & {B-3} & {B-4} & {MET.} & {R-L} & {CIDEr}\\
    \midrule
         dialogue  & .228&.153&.110&.083&.116&.296&0.934\\
         ~ + caption & .229&.154&.111&.084&.119&.305&0.976\\
         ~~~ + vggish & .232&.157&.112&.084&.120&.305&0.994\\
         ~~~~~ + i3d  & \textbf{.239}&\textbf{.161}&\textbf{.116}&\textbf{.087}&\textbf{.121}&\textbf{.309}&\textbf{1.002}\\
    \bottomrule
    \end{tabular}
    \end{center}
    \vspace{-3mm}
    \caption{\label{tab:feature_abalation} Results of the feature ablation tests. Model architectures are simple model using different input feature set. }
    \vspace{-2mm}
    \end{table}

\begin{table}[t!]
    \begin{center}
    \small
    \begin{tabular}{l|ccccccc}
    \toprule
        {Features} & {B-1} & {B-2} & {B-3} & {B-4} & {MET.} & {R-L} & {CIDEr}\\
    \midrule
         vggish & .233&.159&\textbf{.115}&\textbf{.087}&\textbf{.121}&\textbf{.309}&1.019\\
         caption & .232&.158&.114&\textbf{.087}&.120&.307&1.007\\
         dialogue &.234&.159&\textbf{.115}&\textbf{.087}&.120&\textbf{.309}&\textbf{1.023}\\ 
         all & \textbf{.245}&\textbf{.162}&\textbf{.115}&.086&.119&.308&0.977\\ %vggish + caption + dialogue + question &
    \bottomrule
    \end{tabular}
    \end{center}
    \vspace{-3mm}
    \caption{\label{tab:attention_feature} Results of the attention ablation tests. Decoder is attention decoder which attends on different feature set of $\bm{v}$. }
    \vspace{-2mm}
    \end{table}
    
\subsection{Analyzing Fusion Operations}
In Table \ref{tab:fusion_methods}, we test different representation fusion methods to fuse features $\{\bm{z}^1_{T_1}, \dots, \bm{z}^n_{T_n}\}$ into $\bm{v}$, while we choose weighted sum in the simple model (Table~\ref{tab:results}).
3 different fusion methods are listed below:
\begin{compactitem}
    \item Product: $\bm{z} = \prod_i^n \bm{z}^i_{T_i}$
    \item Sum: $\bm{z} = \sum_i^n \bm{z}^i_{T_i}$
    \item Concat: $\bm{z} = W^{cat}[\bm{z}^1_{T_1}; \dots; \bm{z}^n_{T_n}]$, $W^{cat} \in \mathbb{R}^{d_h \times dn}$    
\end{compactitem}
As we expect, the weighted sum outperforms other methods except concat.
While weighted sum is better in CIDEr, concat outperforms it in BLEU scores.
Note that the number of parameters in the weighted sum is significantly smaller than the number in concat, because it only needs to learn $n$ scalars and concat uses $d_hdn$ parameters.
Considering the model size and extensibility, the weighted sum is considered as a better choice.

\subsection{Ablation Test}

In order to investigate the usefulness of features and attention, two sets of experiments are conducted for analysis.

\paragraph{Feature}
In Table \ref{tab:feature_abalation}, we examine the effectiveness of each feature set by fixing the model architecture to the basic model.
The results demonstrate that all feature sets contains semantically meaningful information, and the performance of the basic model gradually improves when we add more features.
When using a simple model architecture, using \textbf{i3d} can improve the performance, which is different from the result shown in Table \ref{tab:results}.
The reason is probably that \textbf{i3d} is too noise to learn, and thus we observe these different results when using \textbf{i3d} features.

\paragraph{Attention}
We also test our attention decoder on different attention values $\bm{v}$. 
Different from our expectation, results in Table \ref{tab:attention_feature} show that different attention values lead to little difference of performance. 
In sum, \textbf{dialogue} gives the best CIDEr score, and using all input features together leads to the best average BLEU score.

\subsection{Official Results}
We submit the following five predictions: 
\begin{compactitem}
    \item Basic fusion model (official): a basic simple fusion model trained on official data
    \item Basic fusion model (prototype): a basic simple fusion model trained on prototype data (w/o and w/ fine-grained response revision)
    \item Fusion text model: a fusion model with multi-stage fusion and attention decoder that only take texts into consideration (w/o and w/ fine-grained response revision)
\end{compactitem}
Note that the submitted models here used different experiment settings from the one proposed in this paper and the models were not well trained.
Here feature selection is not applied, so the basic fusion model considers the \textbf{i3d} features.
The results are shown in Table \ref{tab:official_results}, and the performance is evaluated with consideration of multiple correct responses.
Because the numbers in Table~\ref{tab:official_results} are not comparable with ones in Table~\ref{tab:results}, we also show the submitted results in the last two rows of Table~\ref{tab:results} for fair comparison.
Our simple model outperforms the released baseline in all metrics, especially in CIDEr (by 12\%) and human evaluation (by 11\%). 
The proposed fine-grained response revision brings the small improvement, although the model which is not well trained. 

Our proposed basic fusion architecture has less parameters but achieves better performance than the released baseline, which uses a complicated attention scheme.
The results show that the simple 1$\times$1 convolution fusion may have potential of disentangling each dimension on $z$ shown in Figure \ref{fig:model} and then resulting in more semantically meaningful representations for multiple modalities.

\section{Conclusion}

This paper proposes an intuitive and effective visual dialogue model based on an encoder-decoder design.
We develop a set of modules that are capable of fusing multimodal features with and performing context-aware decoding.
In the experiments of the audio visual scene-aware dialogue task, we validate the effectiveness of each module and analyze whether it is necessary to use all features to correctly answer questions.
The attempt in this paper bridges different modalities and encourages further model exploration for advancing the important research area about multimodality.

\bibliography{ref}

\begin{thebibliography}{}

\bibitem[\protect\citeauthoryear{Alamri \bgroup et al\mbox.\egroup
  }{2018a}]{avsd_dataset}
Alamri, H.; Cartillier, V.; Lopes, R.~G.; Das, A.; Wang, J.; Essa, I.; Batra,
  D.; Parikh, D.; Cherian, A.; Marks, T.~K.; et~al.
\newblock 2018a.
\newblock Audio visual scene-aware dialog (avsd) challenge at dstc7.
\newblock {\em arXiv preprint arXiv:1806.00525}.

\bibitem[\protect\citeauthoryear{Alamri \bgroup et al\mbox.\egroup
  }{2018b}]{alamri2018audio}
Alamri, H.; Cartillier, V.; Lopes, R.~G.; Das, A.; Wang, J.; Essa, I.; Batra,
  D.; Parikh, D.; Cherian, A.; Marks, T.~K.; et~al.
\newblock 2018b.
\newblock Audio visual scene-aware dialog (avsd) challenge at dstc7.
\newblock {\em arXiv preprint arXiv:1806.00525}.

\bibitem[\protect\citeauthoryear{Anderson \bgroup et al\mbox.\egroup
  }{2018}]{anderson2018bottom}
Anderson, P.; He, X.; Buehler, C.; Teney, D.; Johnson, M.; Gould, S.; and
  Zhang, L.
\newblock 2018.
\newblock Bottom-up and top-down attention for image captioning and visual
  question answering.
\newblock In {\em CVPR}, volume~3, ~6.

\bibitem[\protect\citeauthoryear{Antol \bgroup et al\mbox.\egroup }{2015}]{vqa}
Antol, S.; Agrawal, A.; Lu, J.; Mitchell, M.; Batra, D.; Lawrence~Zitnick, C.;
  and Parikh, D.
\newblock 2015.
\newblock Vqa: Visual question answering.
\newblock In {\em Proceedings of the IEEE international conference on computer
  vision},  2425--2433.

\bibitem[\protect\citeauthoryear{{Britz} \bgroup et al\mbox.\egroup
  }{2017}]{Britz:2017}
{Britz}, D.; {Goldie}, A.; {Luong}, T.; and {Le}, Q.
\newblock 2017.
\newblock {Massive Exploration of Neural Machine Translation Architectures}.
\newblock {\em ArXiv e-prints}.

\bibitem[\protect\citeauthoryear{Carreira and
  Zisserman}{2017}]{carreira2017quo}
Carreira, J., and Zisserman, A.
\newblock 2017.
\newblock Quo vadis, action recognition? a new model and the kinetics dataset.
\newblock In {\em Computer Vision and Pattern Recognition (CVPR), 2017 IEEE
  Conference on},  4724--4733.
\newblock IEEE.

\bibitem[\protect\citeauthoryear{Das \bgroup et al\mbox.\egroup
  }{2017}]{visual_dialog}
Das, A.; Kottur, S.; Gupta, K.; Singh, A.; Yadav, D.; Moura, J.~M.; Parikh, D.;
  and Batra, D.
\newblock 2017.
\newblock Visual dialog.
\newblock In {\em Proceedings of the IEEE Conference on Computer Vision and
  Pattern Recognition}, volume~2.

\bibitem[\protect\citeauthoryear{He \bgroup et al\mbox.\egroup
  }{2016}]{he2016deep}
He, K.; Zhang, X.; Ren, S.; and Sun, J.
\newblock 2016.
\newblock Deep residual learning for image recognition.
\newblock In {\em Proceedings of the IEEE conference on computer vision and
  pattern recognition},  770--778.

\bibitem[\protect\citeauthoryear{Hershey \bgroup et al\mbox.\egroup
  }{2017}]{hershey2017cnn}
Hershey, S.; Chaudhuri, S.; Ellis, D.~P.; Gemmeke, J.~F.; Jansen, A.; Moore,
  R.~C.; Plakal, M.; Platt, D.; Saurous, R.~A.; Seybold, B.; et~al.
\newblock 2017.
\newblock Cnn architectures for large-scale audio classification.
\newblock In {\em Acoustics, Speech and Signal Processing (ICASSP), 2017 IEEE
  International Conference on},  131--135.
\newblock IEEE.

\bibitem[\protect\citeauthoryear{Huang \bgroup et al\mbox.\egroup
  }{2018}]{huang2018fusionnet}
Huang, H.-Y.; Zhu, C.; Shen, Y.; and Chen, W.
\newblock 2018.
\newblock Fusionnet: Fusing via fully-aware attention with application to
  machine comprehension.
\newblock In {\em International Conference on Learning Representations}.

\bibitem[\protect\citeauthoryear{Kim \bgroup et al\mbox.\egroup
  }{2016}]{kim2016hadamard}
Kim, J.-H.; On, K.-W.; Lim, W.; Kim, J.; Ha, J.-W.; and Zhang, B.-T.
\newblock 2016.
\newblock Hadamard product for low-rank bilinear pooling.
\newblock {\em arXiv preprint arXiv:1610.04325}.

\bibitem[\protect\citeauthoryear{Kingma and Ba}{2014}]{kingma2014adam}
Kingma, D.~P., and Ba, J.
\newblock 2014.
\newblock Adam: A method for stochastic optimization.
\newblock {\em arXiv preprint arXiv:1412.6980}.

\bibitem[\protect\citeauthoryear{Lin}{2004}]{lin2004rouge}
Lin, C.-Y.
\newblock 2004.
\newblock Rouge: A package for automatic evaluation of summaries.
\newblock {\em Text Summarization Branches Out}.

\bibitem[\protect\citeauthoryear{Luong, Pham, and
  Manning}{2015}]{Luong2015EffectiveAT}
Luong, T.; Pham, H.; and Manning, C.~D.
\newblock 2015.
\newblock Effective approaches to attention-based neural machine translation.
\newblock In {\em EMNLP}.

\bibitem[\protect\citeauthoryear{Papineni \bgroup et al\mbox.\egroup
  }{2002}]{papineni2002bleu}
Papineni, K.; Roukos, S.; Ward, T.; and Zhu, W.-J.
\newblock 2002.
\newblock Bleu: a method for automatic evaluation of machine translation.
\newblock In {\em Proceedings of the 40th annual meeting on association for
  computational linguistics},  311--318.
\newblock Association for Computational Linguistics.

\bibitem[\protect\citeauthoryear{Pennington, Socher, and
  Manning}{2014}]{pennington2014glove}
Pennington, J.; Socher, R.; and Manning, C.~D.
\newblock 2014.
\newblock Glove: Global vectors for word representation.
\newblock In {\em Empirical Methods in Natural Language Processing (EMNLP)},
  1532--1543.

\bibitem[\protect\citeauthoryear{Pirsiavash, Ramanan, and
  Fowlkes}{2009}]{pirsiavash2009bilinear}
Pirsiavash, H.; Ramanan, D.; and Fowlkes, C.~C.
\newblock 2009.
\newblock Bilinear classifiers for visual recognition.
\newblock In {\em Advances in neural information processing systems},
  1482--1490.

\bibitem[\protect\citeauthoryear{Srivastava \bgroup et al\mbox.\egroup
  }{2014}]{JMLR:v15:srivastava14a}
Srivastava, N.; Hinton, G.; Krizhevsky, A.; Sutskever, I.; and Salakhutdinov,
  R.
\newblock 2014.
\newblock Dropout: A simple way to prevent neural networks from overfitting.
\newblock {\em Journal of Machine Learning Research} 15:1929--1958.

\bibitem[\protect\citeauthoryear{Sutskever, Vinyals, and
  Le}{2014}]{sutskever2014sequence}
Sutskever, I.; Vinyals, O.; and Le, Q.~V.
\newblock 2014.
\newblock Sequence to sequence learning with neural networks.
\newblock In {\em Advances in neural information processing systems},
  3104--3112.

\bibitem[\protect\citeauthoryear{Szegedy \bgroup et al\mbox.\egroup
  }{2015}]{szegedy2015going}
Szegedy, C.; Liu, W.; Jia, Y.; Sermanet, P.; Reed, S.; Anguelov, D.; Erhan, D.;
  Vanhoucke, V.; and Rabinovich, A.
\newblock 2015.
\newblock Going deeper with convolutions.
\newblock In {\em Proceedings of the IEEE conference on computer vision and
  pattern recognition},  1--9.

\bibitem[\protect\citeauthoryear{Vaswani \bgroup et al\mbox.\egroup
  }{2017}]{vaswani2017attention}
Vaswani, A.; Shazeer, N.; Parmar, N.; Uszkoreit, J.; Jones, L.; Gomez, A.~N.;
  Kaiser, {\L}.; and Polosukhin, I.
\newblock 2017.
\newblock Attention is all you need.
\newblock In {\em Advances in Neural Information Processing Systems},
  5998--6008.

\bibitem[\protect\citeauthoryear{Vedantam, Zitnick, and
  Parikh}{2015}]{Vedantam2015CIDErCI}
Vedantam, R.; Zitnick, C.~L.; and Parikh, D.
\newblock 2015.
\newblock Cider: Consensus-based image description evaluation.
\newblock {\em 2015 IEEE Conference on Computer Vision and Pattern Recognition
  (CVPR)}  4566--4575.

\bibitem[\protect\citeauthoryear{Wang \bgroup et al\mbox.\egroup
  }{2017}]{P17-1018}
Wang, W.; Yang, N.; Wei, F.; Chang, B.; and Zhou, M.
\newblock 2017.
\newblock Gated self-matching networks for reading comprehension and question
  answering.
\newblock In {\em Proceedings of the 55th Annual Meeting of the Association for
  Computational Linguistics (Volume 1: Long Papers)},  189--198.
\newblock Association for Computational Linguistics.

\bibitem[\protect\citeauthoryear{Xiong, Merity, and
  Socher}{2016}]{xiong2016dynamic}
Xiong, C.; Merity, S.; and Socher, R.
\newblock 2016.
\newblock Dynamic memory networks for visual and textual question answering.
\newblock In {\em International conference on machine learning},  2397--2406.

\end{thebibliography}
\bibliographystyle{aaai}

\begin{table*}[t!]
    \begin{center}
    \small
    \begin{tabular}{c|l|p{11cm}}
    \toprule
        \textbf{Video Frame}&\textbf{Type} & \textbf{Sentences}\\
    \midrule
          \multirow{6}{*}{ \includegraphics[width=1.7cm]{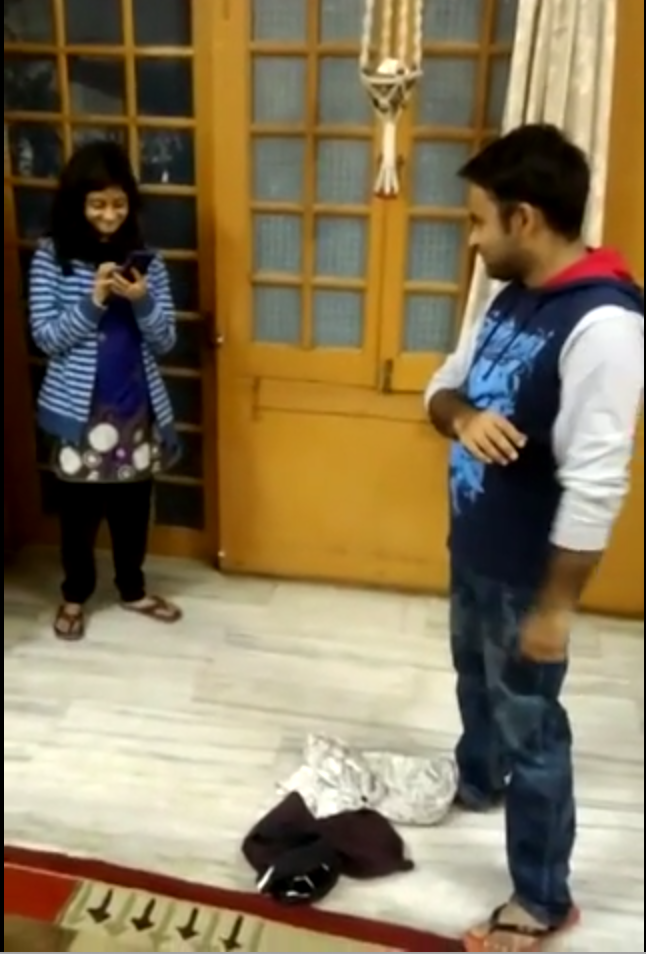}}&Caption & a person in the entryway is working on something on their phone. they start throwing some clothes at another person who is watching them oddly.\\
         &Question & are they laughing in the video? \\
         \cmidrule{2-3}
         &Ground Truth & they are not laughing out loud but are smiling and appear maybe to be flirting a bit.\\
         &Released Baseline & no, they are both talking to each other at the end of the video. \\
         &Basic Fusion Model & no they are not talking. \\
         &Multi-Stage Fusion Model & no, they are not laughing. \\
    \midrule
        \multirow{6}{*}{ \includegraphics[width=1.7cm]{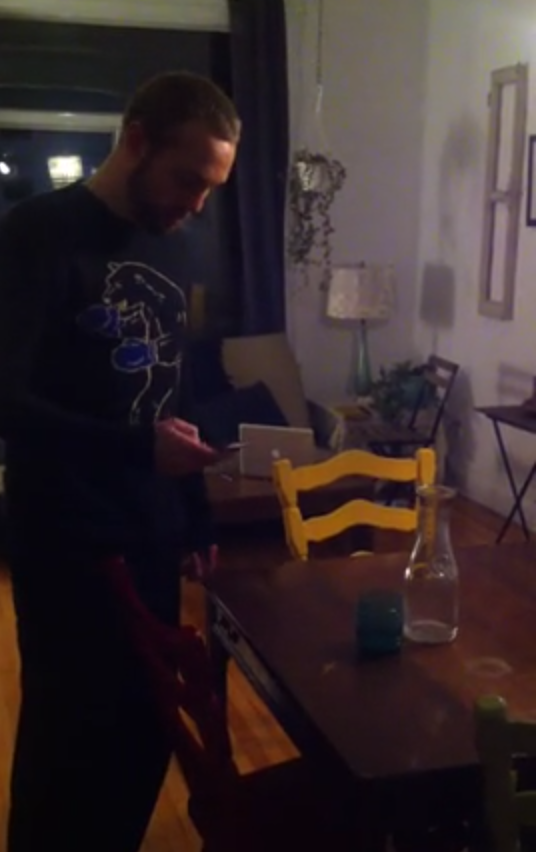}}&Caption & one person walks into the room, pours a glass of something, then grasps a phone and leaves.\\
         &Question & does he appear to be calm in the video?\\
         \cmidrule{2-3}
         &Ground Truth & yes he appears to very calm in the video.\\
         &Released Baseline & no he does not show any emotion. \\
         &Basic Fusion Model & he seems to be in a good mood. \\
         &Multi-Stage Fusion Model & yes he seems to be calm.\\
    \bottomrule
    \end{tabular}
    \vspace{-2mm}
    \end{center}
    \caption{\label{tab:examples} The testing examples of the answers from our models.}
    \end{table*}
    
\newpage
\appendix
    
\section{Appendix}

\subsection{Hyperparameter Setting}

For all text features, we split them by space and use pre-trained word vectors, Glove \cite{pennington2014glove}, to obtain fixed word embedding for each word. 
The hidden state dimension of all uni-direction GRU is 256, and one of  bidirectional GRU is set to 128 for dimension control.
We use the Adam optimizer \cite{kingma2014adam} with $\beta_1 = 0.9$, $\beta_2 = 0.999$ and $\epsilon = 10^{-8}$. 
The initial learning rate is $5 \times 10^{-4}$ and we multiply it by 0.1 every 6,000 update iterations.
We add $10^{-4}$ L2 penalty and apply dropout \cite{JMLR:v15:srivastava14a} to the input word embeddings and video features, where the dropout rate is set to 0.5.
The batch size varies from 8 to 16 according to our available computing resources as regularization.
During training, the early stop mechanism is applied.
%Each model is trained to converge and we select the checkpoint with best validation loss. 
During testing, we use beam search and set the beam size equal to 5 to generate our final answers.

\subsection{Qualitative Analysis}

We show some examples in Table \ref{tab:examples} to qualitatively evaluate our model.
In the first example, a woman is using her phone, while a man is staring at her in the video. 
One question in the dialogue about this video is ``{\it Are they laughing in the video?}'', which should be easy for the model with incorporation of audio features (\textbf{vggish}). 
However, the released baseline says ``{\it no, they are both talking to each other at the end of the video}'', which is totally wrong because they only appear to be happy and a little flirting but do not talk or laugh.
On the other hand, our model is able to correctly capture the audio features and answer they don't talk and laugh in the video.
An interesting difference of the basic fusion model and the multi-stage fusion model is the word they choose.
The basic fusion model uses ``\emph{talking}'' while the multi-stage model picks the word ``\emph{laughing}'', which is more proper for this question.
This example demonstrates that our proposed model can help system understand and catch what happens in the video and what the questioner asks. 

In the second example, a person walks into the room, drink a water, picks a phone and leaves.
It is difficult to tell whether the man is calm from few video frames, because face expression of the man is not clear in the video.
However, our model can answer correctly using the fact that this man does not make large sound, which is the evidence that this man moves slowly, and calmly. 
The previous rounds of the dialogue mention that this man pour water to the coup and drink from it, which is another evidence that this man is not hurry to pick his phone and leave. 
To answer this hard question correctly, it is necessary to fuse evidences from both audio and texts, and our model is capable of generating the correct answer by fusing the multimodal features properly.

\end{document}